\documentclass[10pt,twocolumn,letterpaper]{article}

\usepackage{cvpr}
\usepackage{times}
\usepackage{epsfig}
\usepackage{graphicx}
\usepackage{amsmath}
\usepackage{amssymb}

\usepackage{color}

\usepackage{subfigure}

\usepackage{interval}

\usepackage{caption}

\usepackage{pifont}

\usepackage{makecell}
\usepackage{booktabs}
\usepackage{multirow}
\usepackage{siunitx}

\usepackage[T1]{fontenc}
\usepackage{fix-cm}

\newcolumntype{?}{!{\vrule width 1pt}}
\newcolumntype{C}[1]{>{\centering}m{#1}}

\newcolumntype{X}{@{\hskip\tabcolsep\vrule width 1.5pt\hskip\tabcolsep}}

\usepackage[pagebackref=true,breaklinks=true,letterpaper=true,colorlinks,bookmarks=false]{hyperref}

\cvprfinalcopy 

\def\cvprPaperID{2236} 

\ifcvprfinal\fi
\begin{document}

\title{Egocentric Basketball Motion Planning \\ from a Single First-Person Image}

\author{Gedas Bertasius\\
University of Pennsylvania\\
{\tt\small gberta@seas.upenn.edu}
\and
Aaron Chan\\
University of Southern California\\
{\tt\small chanaaro@usc.edu}
\and
Jianbo Shi\\
University of Pennsylvania\\
{\tt\small jshi@seas.upenn.edu}
\and
}

\maketitle

\begin{abstract}



We present a model that uses a single first-person image to generate an egocentric basketball motion sequence in the form of a 12D camera configuration trajectory, which encodes a player's 3D location and 3D head orientation throughout the sequence. To do this, we first introduce a future convolutional neural network (CNN) that predicts an initial sequence of 12D camera configurations, aiming to capture how real players move during a one-on-one basketball game. We also introduce a goal verifier network, which is trained to verify that a given camera configuration is consistent with the final goals of real one-on-one basketball players. Next, we propose an inverse synthesis procedure to synthesize a refined sequence of 12D camera configurations that (1) sufficiently matches the initial configurations predicted by the future CNN, while (2) maximizing the output of the goal verifier network. Finally, by following the trajectory resulting from the refined camera configuration sequence, we obtain the complete 12D motion sequence.

Our model generates realistic basketball motion sequences that capture the goals of real players, outperforming standard deep learning approaches such as recurrent neural networks (RNNs), long short-term memory networks (LSTMs), and generative adversarial networks (GANs).

\end{abstract}

\vspace{-0.5cm}

\section{Introduction}

Consider LeBron James, arguably the greatest basketball player in the world today. People often speculate about which of his traits contributes most to his success as a basketball player. Some may point to his extraordinary athleticism, while others may credit his polished basketball mechanics (i.e., his ability to accurately pass and shoot the ball). While both of these characteristics are certainly important, there seems to be a general consensus, among sports pundits and casual fans alike, that what makes LeBron truly exceptional is his ability to make the right decision at the right time in almost any  basketball situation.


\captionsetup{labelformat=default}
\captionsetup[figure]{skip=10pt}

\begin{figure}
\begin{center}
   \includegraphics[width=0.92\linewidth]{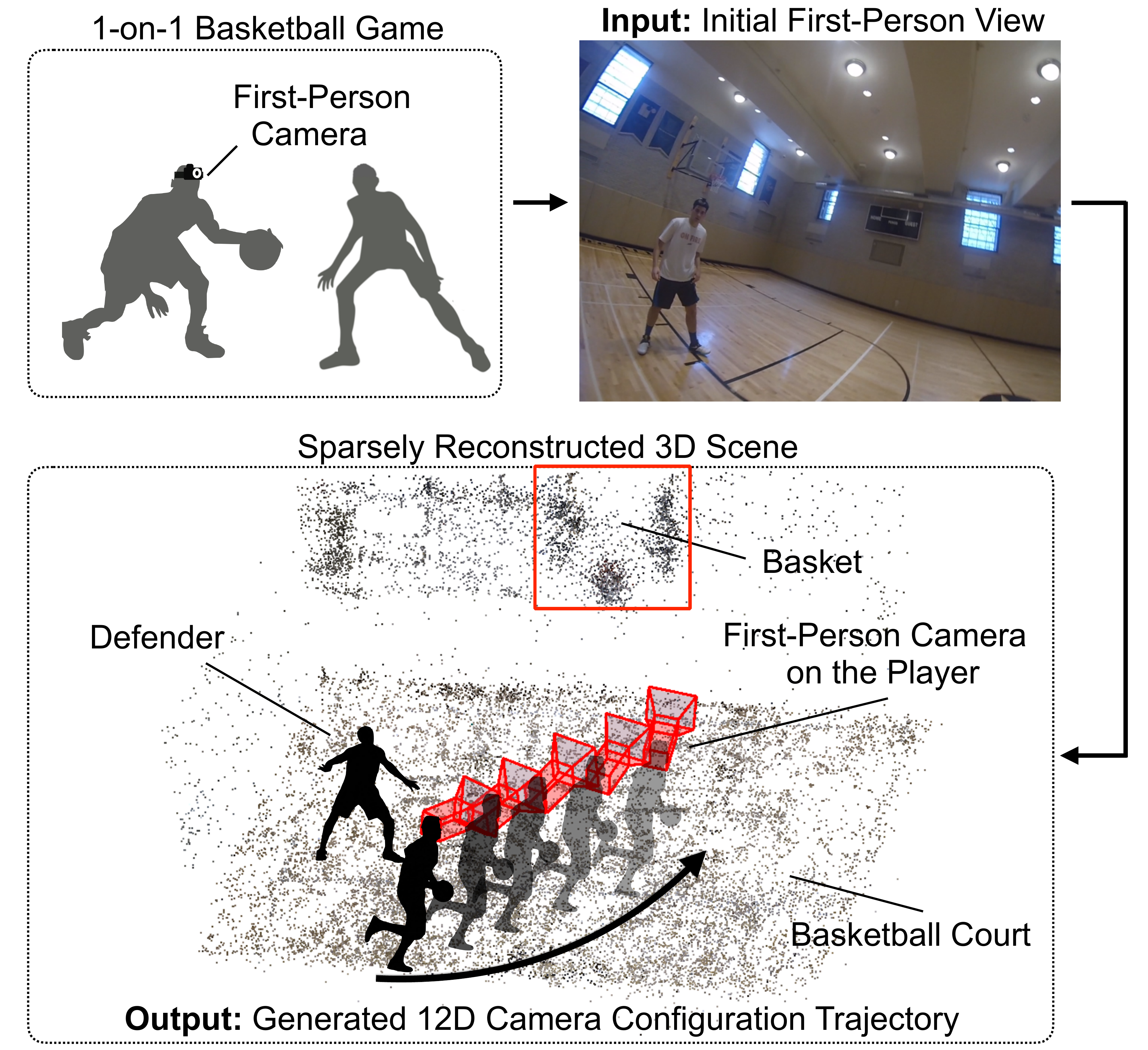}
\end{center}
\vspace{-0.4cm}
   \caption{Given a single first-person image from a one-on-one basketball game, we aim to generate an egocentric basketball motion sequence in the form of a 12D first-person camera configuration trajectory, which encodes a player's 3D location and 3D head orientation. In this example, we visualize our generated motion sequence within a sparse 3D reconstruction of an indoor basketball court. \vspace{-0.5cm}}
   
\label{main_fig}
\end{figure}


Now, imagine yourself inside the dynamic scene shown in Figure~\ref{main_fig}, where, as a basketball player, you need to make a series of moves to outmaneuver your defender and score a basket. In doing so, you must evaluate your position on the court, your defender's stance, your defender's court position relative to the basket, and many other factors, if you want to maximize your probability of scoring. For instance, if you observe that your defender is positioned close to you and leaning more heavily on \textit{his} right leg, you might exploit this situation by taking a swift left jab-step followed by a hard right drive to the basket, throwing your defender off balance and earning you an easy layup. However, if your defender is standing farther away from you, you may decide to take a few dribbles into an area where your defender cannot get to you quickly, before stopping for a jump-shot. These are all complex decisions that have to be made within a split second -- doing this consistently well is challenging. 

In this paper, we aim to build a model that mimics this basketball decision-making process.  Specifically, our model maps a first-person visual signal to a plausible egocentric basketball motion sequence. This is a difficult problem because little is known about how skilled players make subsecond-level decisions. What we do know is that players make decisions based on what they see: a player may look at how his or her defender is positioned (i.e., feet orientation, torso orientation, etc.) and then, based on that visual information, choose to drive right or left. Leveraging this insight, in this work, we learn our model from data recorded using first-person cameras. First-person cameras allow us to see various subtle details in the basketball game, just as the players do. In contrast, a standard third-person camera typically records a low-resolution view from a suboptimal orientation, making it difficult to observe such details.  Because these subtle details may be crucial for predicting where a player will move next, first-person modeling is beneficial.

Our model takes a single first-person image as input and generates an egocentric basketball motion sequence in the form of a 12D first-person camera configurations, encoding a player's 3D location and 3D head orientation throughout the sequence. To do this, we first introduce a future convolutional neural network (CNN) that predicts an initial sequence of 12D camera configurations, aiming to capture how real players move during a one-on-one basketball game. We also introduce a goal verifier network, which is trained to verify that a given camera configuration is consistent with the final goals of real players. Next, we use our proposed inverse synthesis procedure to synthesize a refined sequence of 12D camera configurations by optimizing the following objectives: (1) minimize the difference between the refined configurations and initial configurations predicted by future CNN while also (2) maximizing the goal verifier network output. Finally, by following the trajectory resulting from the refined camera configuration sequence, we obtain the complete 12D motion sequence.

Our egocentric basketball motion model learns to generate smooth and realistic sequences that capture the goals of real basketball players. Additionally, our model is learned in an unsupervised fashion; this is an important advantage, since obtaining labeled behavior data is costly. Finally, in our experimental section, we show that our method consistently outperforms standard deep learning approaches such as RNNs~\cite{Jain:1999:RNN:553011}, LSTMs~\cite{Hochreiter:1997:LSM:1246443.1246450}, and GANs~\cite{NIPS2014_5423}.

\section{Related Work}

\textbf{Using Vision for Behavior Modeling.} Developing models that characterize human behavior in everyday tasks, such as walking or driving, has been a long-standing problem in computer science. The work in~\cite{Pentland:1999:MPH:307835.307862} uses a hidden Markov model (HMM) to learn human driving patterns. Recent work in~\cite{Liu2017ImitationFO,DBLP:journals/corr/RahmatizadehABL17} uses third-person videos of humans performing simple tasks, like opening a door, to teach robots how to do the same. By observing them from a third-person view, the method in~\cite{chenxiawu_watchnbots_icra2016} learns to remind humans of actions they forgot to do. Recently, there also has been a surge of methods designed to capture various aspects of social human behavior. For instance, the work in~\cite{Ali2008} uses a flow field model to track crowds. Also, the work in~\cite{conf/iccv/PellegriniESG09} develops a tracking method that can predict walking trajectories for multiple agents simultaneously, while more recent work in~\cite{Ma_2017_CVPR} predicts walking trajectories by using a game theoretic approach. Furthermore, the methods in~\cite{PhysRevE.51.4282,conf/cvpr/MehranOS09,conf/cvpr/AlahiRF14,dasgupta_cvpr16} develop social force or social affinity based methods for predicting pedestrian behaviors. In addition, the method in~\cite{lee2016predicting}, uses Markov decision processes to predict the wide receiver trajectories in football, whereas the work in~\cite{soccer_ghosting} learns to predict the behavior of soccer players. Finally, the work in~\cite{NIPS2016_6520} uses large amounts of ``top-view'' basketball tracking data~\cite{stats_vu} to train deep hierarchical networks for basketball trajectory prediction.



In contrast to these prior methods, our data are obtained via first-person cameras, which allows us to build a model that connects first-person visual sensation to egocentric motion planning ability. Using such first-person data is our main advantage compared to prior methods.




   \begin{figure*}
\begin{center}
   \includegraphics[width=0.95\linewidth]{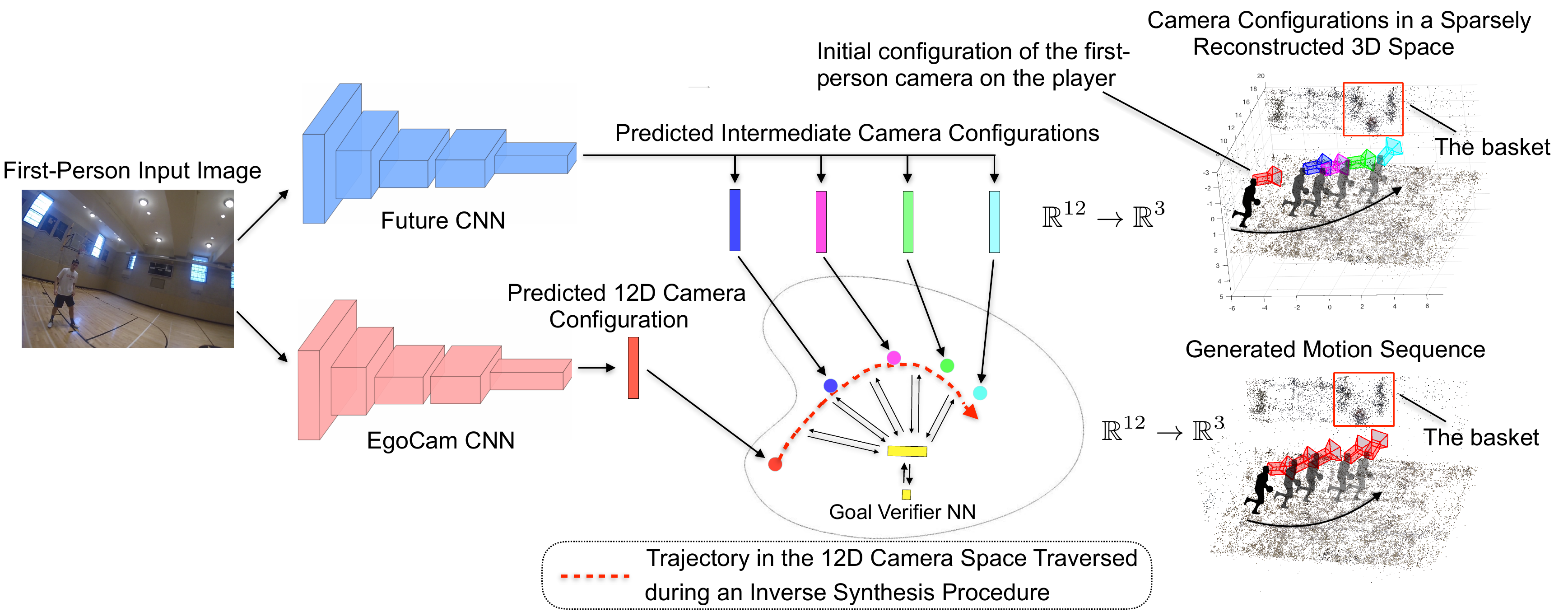}
\end{center}
\vspace{-0.6cm}
\caption{Our model takes a single first-person image as input and outputs an egocentric basketball motion sequence in the form of a 12D camera configuration trajectory, encoding a player's 3D location and 3D head orientation throughout the sequence. First, we feed the first-person image through our proposed future CNN to predict an initial sequence of future 12D camera configurations. Then, we use our proposed inverse synthesis procedure to synthesize a refined camera configuration sequence that matches the configurations predicted by the future CNN, while also maximizing the output of the goal verifier network. The goal verifier network is a fully-connected network trained to verify that a given 12D camera configuration is consistent with the final goals of real players. Finally, by following the trajectory resulting from the refined camera configuration sequence, we obtain the complete 12D motion sequence. \vspace{-0.4cm}}
\label{arch_fig}
\end{figure*}

\textbf{First-Person Vision.}  Most first-person methods have focused on first-person object detection~\cite{DBLP:journals/ijcv/LeeG15,BMVC.28.30,conf/cvpr/RenG10,conf/cvpr/FathiRR11,gberta_2017_RSS,gberta_2017_ICCV_vsn} or activity recognition~\cite{conf/cvpr/KitaniOSS11,Soran2015,Singh_2016_CVPR,PirsiavashR_CVPR_2012_1,Li_2015_CVPR,ma2016going,Fathi:2011:UEA:2355573.2356302}. Several methods have also employed first-person videos for video summarization ~\cite{DBLP:journals/ijcv/LeeG15,Lu:2013:SSE:2514950.2516026}. Furthermore, recent work in~\cite{Su2016} introduced a first-person method for predicting the camera wearer's engagement, while ~\cite{gberta_2017_ICCV_baller} developed a model to assess a basketball performance from a first-person video.  Additionally, ~\cite{Rhinehart_2017_ICCV} proposed applying inverse reinforcement learning on first-person data to infer the goals of the camera wearer during daily tasks. Furthermore, the methods in~\cite{park_ego_future,park_cvpr:2017} propose to generate plausible walking trajectories from first-person images. However, the methods in~\cite{park_cvpr:2017,park_ego_future} generate walking trajectories without really understanding what the camera-wearer's actual goals are, often resulting in overly simplistic and unrealistic trajectories.


In contrast to these prior methods, we consider a complex one-on-one basketball game scenario, for which we not only infer the goals of the camera-wearer, but also generate an egocentric basketball motion sequence that is aligned with these goals. We also point out that, unlike the methods in~\cite{park_cvpr:2017,park_ego_future} which require depth input or the positions of other people in the scene, our method operates using only a single first-person RGB input image. 

\section{Motivation and Challenges}

\textbf{Representing a State.}  Generating an egocentric basketball motion sequence can be viewed as the problem of mapping a first-person image $x$ to a sequence $y_{1:M}$, where each entry $y_i$ is a vector representing a state in this sequence of length $M$. In this paper, we use a notation where the entire sequence is denoted as $y$ (without subscript indices), whereas an $i^{th}$ state in the sequence $y$ is denoted as $y_i$.


We propose to encode each state $y_i$ in the sequence as a 12D camera configuration, which captures 3D location and 3D orientation of the camera on a player's head.  In contrast to prior models~\cite{park_ego_future,park_cvpr:2017} that just use a 2D $(x,y)$ location representation, our selected camera configuration representation allows us to represent more complex basketball motion patterns, while still being compact and interpretable. Such representation also enables us to learn our model without using manually labeled behavioral data.

\textbf{Generating Motion Sequences.} Generating motion sequences that are realistic and smooth is a challenging problem because the actions taken by the camera wearer are noisy. As a result, it is difficult to accurately learn the transitions between two adjacent states from a limited amount of data. Furthermore, due to the recurrent nature of such predictions (i.e., the predicted state $\hat{y}_{i+1}$ depends on $\hat{y}_{i}$), the error starts accumulating and exploding, which makes it difficult to generate longer sequences accurately. This issue is especially prevalent with the standard RNN and LSTM approaches, which try to learn such transitions sequentially. 

In this work, we propose a model that first predicts a set of intermediate states that are consistent with how real basketball players move. Afterwards, we use our proposed inverse synthesis procedure to generate a full motion sequence. In the experimental section, we verify the effectiveness of our approach and show that it outperforms methods such as RNNs, LSTMs, and GANs.





\section{Egocentric Basketball Motion Model}


In Figure~\ref{arch_fig}, we present a detailed illustration of our egocentric basketball motion model. First, we use our proposed future CNN to predict an initial sequence of 12D camera configurations, which aim to capture how real players move during a one-on-one basketball game. After that, we use our proposed inverse synthesis procedure to generate a refined camera configuration sequence that (1) matches the initial camera configuration sequence predicted by the future CNN and (2) maximizes the output of the goal verifier network, which is trained to verify whether a given 12D camera configuration aligns with realistic one-on-one basketball goals. Finally, by following the trajectory resulting from the refined camera configuration sequence, we obtain the complete 12D motion sequence. We now discuss each component of our model in more detail. 


   \begin{figure*}
\begin{center}
   \includegraphics[width=0.9\linewidth]{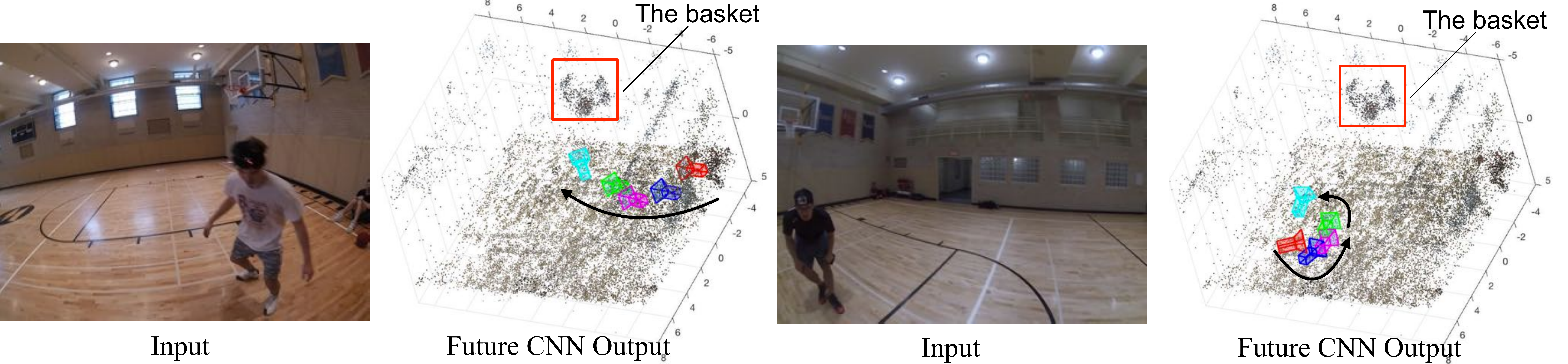}
\end{center}
\vspace{-0.6cm}
\caption{An illustration of the camera configurations generated by our future CNN, which we visualize in a sparsely reconstructed 3D space (best viewed in color). The \text{\color{red} red} camera depicts the initial camera configuration state, while the \text{\color{blue} blue}, \text{\color{magenta} magenta}, \text{\color{green} green}, and \text{\color{cyan} cyan} cameras correspond to the outputs from the $1^{st}, 2^{nd}, 3^{rd}$, and $4^{th}$ branches of our future CNN, respectively. We note that our future CNN is able to produce a diverse set of intermediate configurations, which allows us to generate a wide array of different sequences at a later step in our model. \vspace{-0.4cm}}

\label{fut_vis_fig}
\end{figure*}

\subsection{Egocentric Camera (EgoCam) CNN}

An egocentric camera (EgoCam) CNN is used to map a given first-person image $x_i$ into a 12D camera configuration $\hat{y}_i \in \mathbb{R}^{1 \times 12}$ that encodes the 3D location and 3D orientation of the camera on a player's head. We implement our EgoCam CNN using a popular ResNet-101~\cite{DBLP:journals/corr/HeZRS15} architecture. The network is optimized using the following L2 loss:

\vspace{-0.3cm}
\begin{equation}
\begin{split}
L_{cam} & = || \hat{y}_i(x_i) - K(x_i) ||^2_2\\
\end{split}
\end{equation}

where $\hat{y}_i \in \mathbb{R}^{1 \times 12}$ is the predicted camera configuration for an image $x_i$, and $K(x_i) \in \mathbb{R}^{1 \times 12}$ is a vector that encodes a true player's 3D location and 3D head orientation. $K(x_i)$ is obtained by flattening the egocentric $3 \times 4$ camera matrix, which is produced by an unsupervised structure from motion (SfM) algorithm. However, due to the jittery nature of first-person images, SfM works only on a small portion of the input images (i.e.$ \approx 20\%$ of all images). This motivates the need for an EgoCam CNN, which allows us to generate 12D configurations for any first-person image. 

Our EgoCam CNN produces similar outputs to the actual SfM algorithm. We also report that we tried retrieving missing 12D configurations using a nearest neighbor algorithm, but observed that our EgoCam CNN produced better results (\textbf{2.54} vs. \textbf{1.97} L2 error in the normalized 12D space).  



\subsection{Future CNN}



Given a first person image $x_i$ from the beginning of a sequence, the future CNN encodes it into multiple configurations $\phi(x_i) \in \mathbb{R}^{k \times 12}$, which represent intermediate states that capture how real players move during a one-on-one game. Here, $k$ is a parameter that controls how many intermediate configurations we want to generate. Our future CNN is implemented using a multi-path ResNet-101 architecture, meaning that after the final convolutional $\textit{pool5}$ layer, the network splits itself into $k$ branches, each denoted as $\phi_j(x_i)$ for $j=1 \hdots k$. Each branch in the future CNN is responsible for generating its own intermediate state. 

To train the future CNN, for every real basketball sequence, we first assign to each camera configuration $y_i$ of the sequence a value $s \in [0,1]$, indicating its order in the sequence. Let $y_i$ be the $i^{th}$ configuration in a sequence of length $M$. Then, we can compute $s$ as $s(i)=i/M$. The $j^{th}$ branch in the future CNN is then optimized to predict camera configurations, with $s$ values falling in the interval $[\frac{j-1}{k},\frac{j}{k}]$, where $k$ is the number of branches in the future CNN. For instance, if $k=4$, then the second branch in the network will be optimized to predict camera configurations with values $s \in [0.25,0.5]$. This constraint ensures that each branch generates an intermediate state associated with a specific time-point in a sequence, thus generating configurations covering the entire sequence (albeit with wider gaps). We also point out that all input images $x_i$ that are used to train the future CNN have values $s \in [0,0.1]$ to make sure that the future CNN generates accurate intermediate configurations from a first-person image at the beginning of a sequence. The future CNN is then trained using the following loss function:


\vspace{-0.3cm}
\begin{equation}
\begin{split}
 & L_{fut} =  \sum_{j=1}^k || \phi_j(x_i) - \hat{y}_{i'}(x_{i'}) ||^2_2\\ 
\end{split}
\end{equation}

where $\phi_j$ denotes the output from branch $j$ of a future CNN, and $\hat{y}_{i'}$ is the output of an EgoCam CNN for an input image $x_{i'}$. The constraints on $x_i$ and $x_{i'}$ that we discussed above are expressed as: $s(i) \in [0, 0.1]$ and  $s(i') \in [\frac{j-1}{k},\frac{j}{k}]$. 

We note that our training setup requires basketball sequences that are trimmed. In this work, we trim the sequences manually: the sequence starts when a player begins his offensive position and ends when a shot or a turnover occurs.  We believe that we could also trim sequences automatically by using a player's head location and pose to assess whether a player is attacking or defending.

 \begin{figure*}
\begin{center}
   \includegraphics[width=0.84\linewidth]{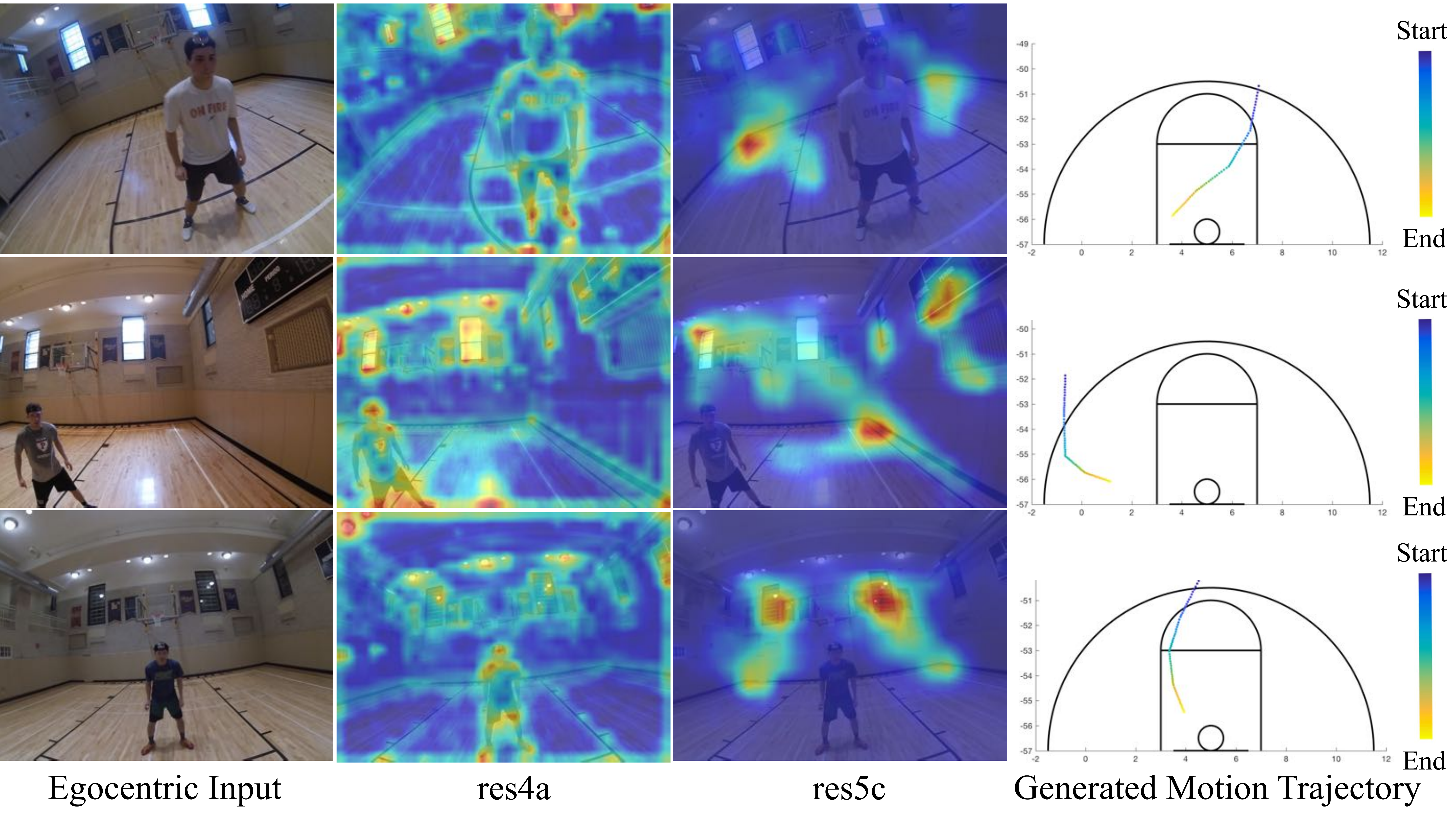}
\end{center}
\vspace{-0.6cm}
\caption{A figure illustrating some of our qualitative results. In the first column, we depict an egocentric input image. In the second and third columns, we visualize the activations of a Future CNN from the res4a and res5c layers. Finally, in the last column, we present our generated 2D motion trajectories. Based on the activations of the Future CNN, we conclude that our CNN recognizes visual cues in an egocentric image, which may be helpful for deciding how to effectively navigate the court and reach the basket, or how to get away from a defender.  Furthermore, our generated motion trajectories seem realistic, as they avoid colliding with the defender and typically end near the basket, which reflects how most real players would move.\vspace{-0.4cm}}

\label{cam_seqs_fig}
\end{figure*}

\subsection{Goal Verifier Network}

The goal verifier network aims to verify whether a given configuration aligns with the final goals of real players. The goal verifier takes a 12D camera configuration $\hat{y}_i \in \mathbb{R}^{1 \times 12}$ as input, then outputs a real value $\psi(\hat{y}_i)  \in \mathbb{R}^{1 \times 1}$ in the interval $[0,1]$, indicating how well a given camera configuration captures the final goals of real players. 

The key question is: how do we infer the final goals of real basketball players without asking them directly? To do this, we assume that the last few images in a given sequence represent the final goals of that player. Of course, such reasoning may not always be correct because, in some sequences, players may fail to accomplish their goals. However, we conjecture that when looking at many sequences of real players, the pattern of goals should be quite distinctive. In other words, if we use the right learning strategy, the network should be able to learn configurations that are typically associated with the final goals of real players. 

We propose to employ discriminative training of the goal verifier, which allows us to effectively address the issues of noisy labels in our setting. To create these noisy labels, we assign a value $g(\hat{y}_i)=1$ to all configurations with $s(i)>0.92$ (configurations that appear at the end of sequences) and a value of zero to all other configurations. Such a scheme allows us to highlight the configurations at the end of sequences, which are more likely to represent the goals of the players. Subsequently, the goal verifier, which is a two-layer network, takes a 12D camera configuration as its input and is trained using a cross-entropy loss:




\vspace{-0.3cm}
\begin{equation}
\begin{split}
L_{g} = -\big[ g(\hat{y}_i) \log{\psi(\hat{y}_i})+(1-g(\hat{y}_i)) \log \left(1- \psi(\hat{y}_i\right)) \big]
\end{split}
\end{equation}

where $\hat{y}_i \in \mathbb{R}^{1 \times 12}$ is the 12D camera configuration output from an EgoCam CNN (associated with an image $x_i$), and $\psi(\hat{y}_i)  \in \mathbb{R}^{1 \times 1}$ is the output of a goal verifier network that indicates whether a given 12D camera configuration accurately represents the final goals of real players. 

\subsection{Motion Planning using Inverse Synthesis}

We can now put all the pieces of our model together and show how to generate an egocentric basketball motion sequence that captures the goals of real players. Our approach is partially inspired by the idea of synthesizing the preferred stimuli in a CNN~\cite{DBLP:journals/corr/YosinskiCNFL15,Mahendran15,Mahendran16}, which has been mostly used to visualize activations in the CNNs~\cite{DBLP:journals/corr/YosinskiCNFL15,Mahendran15,Mahendran16}. In this work, we propose the concept of inverse synthesis for generating a realistic basketball motion sequence.

Consider the two problems that we were solving previously: (1) encoding an egocentric basketball motion sequence via a set of intermediate 12D configurations and (2) verifying that a given 12D camera configuration aligns with the final goals of real players. In the previous sections, we solved these two problems using the future CNN and the goal verifier network, respectively. However, we now want to invert these two problems: given (1) a set of intermediate states predicted by the future CNN and (2) a trained goal verifier network, our goal is to generate a smooth basketball motion sequence in the form of 12D camera configurations. 

To do this, we frame the inverse synthesis problem as a search problem in the 12D camera space, where our goal is to find a 12D configuration $h$ that (1) matches intermediate states generated by the future CNN $\phi_j(x_i)$ and (2) maximizes the output of a goal verifier network $\psi(h)$. We formulate this problem as a minimization problem in the 12D camera configuration space:

\vspace{-0.3cm}
\begin{equation}
\begin{split}
h^* =\arg\min_{h} \big[ || \phi_j(x_i) - h ||^2_2 - \log{(\psi(h))} \big]
\end{split}
\end{equation}

\begin{figure}
\begin{center}
   \includegraphics[width=0.5\linewidth]{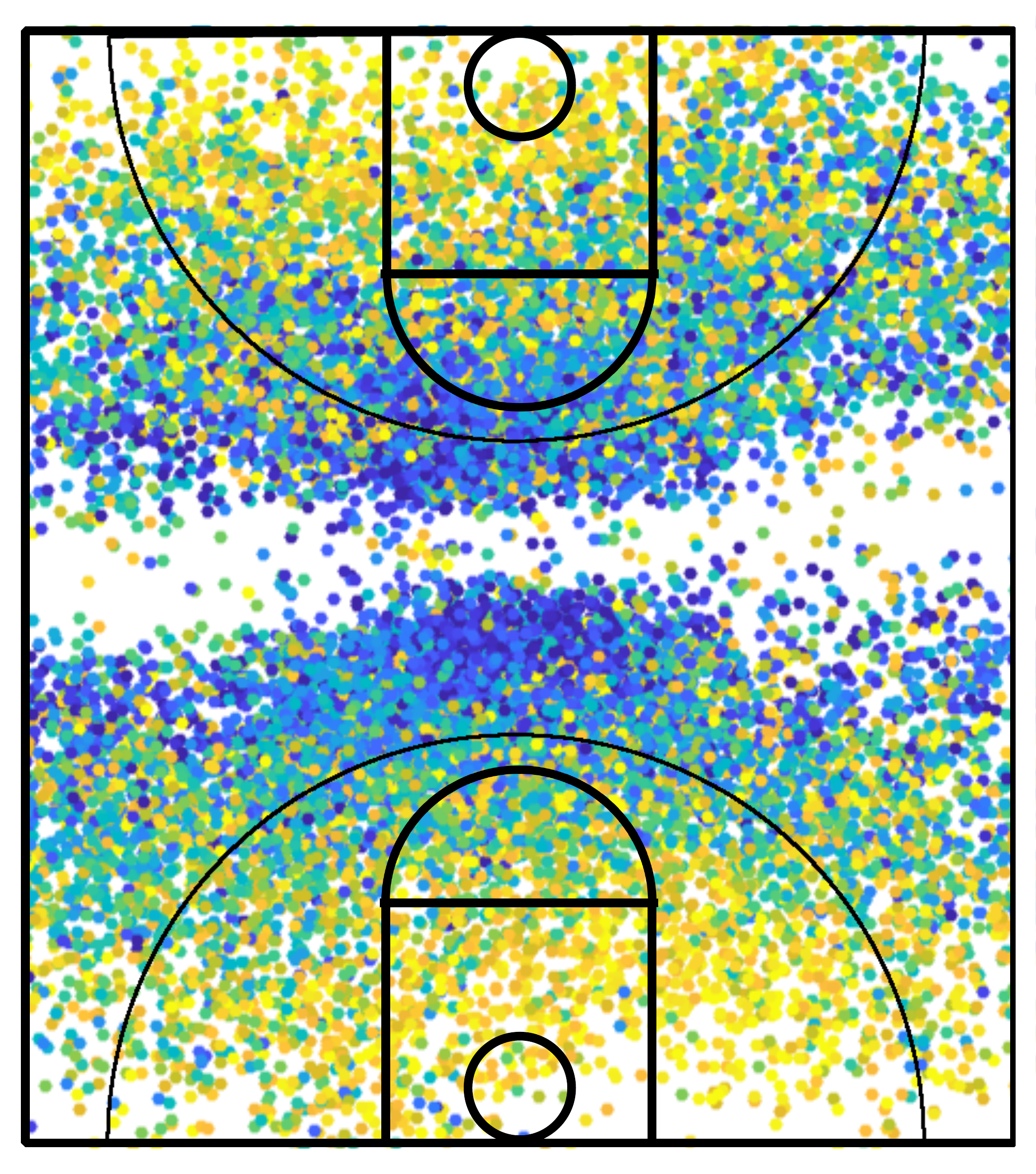}
\end{center}
\vspace{-0.4cm}
   \caption{A figure illustrating the 3D locations of every camera configuration from our dataset mapped on a 2D court (best viewed in color). The blue points indicate the configurations from the beginning of sequences, whereas the yellow points depict configurations from the end of sequences. The diversity of real sequences in our dataset allows us to build a powerful basketball motion model.\vspace{-0.5cm}}
\label{data_fig}
\end{figure}

Here, $h$ is a 12D camera configuration vector that we initialize to $\hat{y}_i(x_i)$, which is the camera configuration of the very first image in the sequence (i.e., $s(i)=0$, or, equivalently, $i=0$).  The first term in the equation encourages $h$ to reach configurations predicted by the future CNN, whereas the second term encourages $h$ to take values that would maximize the output of the goal verifier network. We minimize this function by computing the appropriate gradients and then running gradient descent in the 12D camera configuration space for $N$ iterations. To select the branch from a future CNN whose output $\phi_j$ we are trying to match, we compute $j=floor(c/(N/k))$, where $j$ is the index of a selected branch, $c$ is the iteration counter, and $k$ is the number of branches in the future CNN. 

During the inverse synthesis procedure, we fix the parameters of all three networks and only adjust $h$. Since the inverse synthesis is performed in the 12D camera configuration space, at every step we are generating a new 12D camera configuration. Thus, at the end, we can generate a final camera configuration sequence by simply following the 12D camera trajectory traversed during the optimization.

\subsection{Implementation Details}

For all of our experiments, we used the Caffe library~\cite{jia2014caffe}. Both the future and EgoCam CNNs were based on the ResNet-101~\cite{DBLP:journals/corr/HeZRS15} architecture and were trained for $10,000$ iterations, with a learning rate of $10^{-4}$, $0.9$ momentum, a weight decay of $5 \cdot 10^{-4}$, and $20$ samples per batch. We designed our future CNN to have $4$ distinct branches, which we experimentally discovered to work well. Each branch consists of a single fully-connected layer that maps $\textit{pool5}$ features to a 12D camera configuration feature. Next, we designed the goal verifier network as a two layer fully connected network with $100$ neurons in the first-hidden layer and a single output neuron at the end. To generate final basketball motion sequences, we ran our inverse synthesis procedure for $6,000$ iterations with a step size of $0.001$.


 \setlength{\tabcolsep}{3.5pt}

   \begin{table}[t]
   \small
    \begin{center}
    \begin{tabular}{ c  | c |  c |}
    \cline{2-3}
    & \multicolumn{2}{ c |}{Evaluation Tasks} \\
    \cline{2-3}
      &   \multicolumn{1}{ c |}{PFM $\downarrow$}  &  \multicolumn{1}{ c |}{CG $\uparrow$} \\ \cline{1-3}
      \multicolumn{1}{| c |}{GAN~\cite{NIPS2014_5423}} & 62.31 & 0.329\\
       \multicolumn{1}{| c |}{LSTM~\cite{Hochreiter:1997:LSM:1246443.1246450}} &  5.66 & 0.678 \\
       \multicolumn{1}{| c |}{RNN~\cite{Jain:1999:RNN:553011}} & 5.82 & 0.612\\
      \multicolumn{1}{| c |}{NN}  &  5.36 & -\\ \hline
       \multicolumn{1}{| c |}{\bf Ours w/o GV}  & \bf 4.91 & 0.671\\ 
       \multicolumn{1}{| c |}{\bf Ours w/ GV}  &  4.93 & \bf 0.776\\ \hline
       
    \end{tabular}
    \end{center}
    \vspace{-0.2cm}
    \caption{We assess each method on two tasks: predicting a player's future motion (PFM), and capturing the goals of real players (CG). The $\uparrow$ and $\downarrow$  symbols next to a task indicate whether a higher or lower number is better, respectively. Our method outperforms the baselines for both tasks, suggesting that we can generate sequences that (1) are more similar to a true player's motion and (2) capture the goals of real players more accurately. We also note that removing a goal verifier (``Ours w/o GV") leads to a sharp decrease in performance, according to the CG evaluation metric. \vspace{-0.4cm}} 
    \label{quant_table}
   \end{table}

   \setlength{\tabcolsep}{1.5pt}

\captionsetup{labelformat=default}
\captionsetup[figure]{skip=10pt}

\begin{figure*}
\begin{center}
    \includegraphics[width=0.95\linewidth]{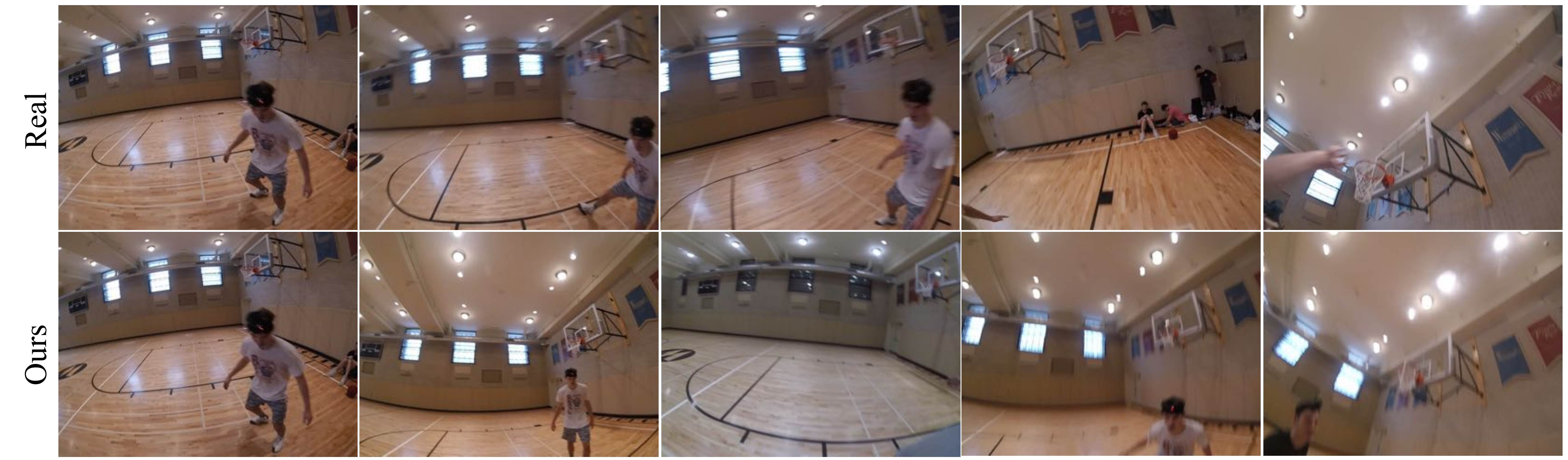}
\end{center}
\vspace{-0.4cm}
   \caption{A visual illustration of our generated basketball sequences, in the form of first-person images retrieved using nearest neighbor search. We observe that our generated sequence is similar to a real sequence, suggesting that our model could be used for applications such as future behavior prediction.\vspace{-0.4cm}}
   
\label{img_fig}
\end{figure*}

\section{Egocentric One-on-One Basketball Dataset}

We present a first-person basketball dataset consisting of $988$ sequences from one-on-one basketball games between nine college-level players. The videos are recorded in 1280$\times$960 resolution at 100 fps using GoPro Hero Session cameras, which are mounted on the players' heads. We extract the videos at $5$ frames per second. We then randomly split all the sequences into training and testing sets, consisting of $815$ and $173$ sequences, respectively. Each sequence is about $25$ frames long. 

To obtain better insight about the distribution of our data, we map the 3D locations of all camera configurations from real sequences onto the 2D court and visualize the results in Figure~\ref{data_fig}. The blue points indicate the configurations at the beginning of sequences, whereas the yellow points represent configurations at the end of sequences. Based on this figure, we note that our dataset contains a wide array of different beginning and ending configurations, enabling us to learn a diverse basketball motion model.


We chose a one-on-one setting because it is more data-efficient for studying the decision-making process of basketball players (compared to the five-on-five setting). For example, first-person videos of five-on-five games capture numerous instances when players do not have the ball and are idly waiting for their teammate with the ball to perform some action. On the contrary, in our one-on-one dataset, players continuously make decisions and take actions to reach their goals, as there are no teammates to rely on. 


\section{Experimental Results}

To the best of our knowledge, we are the first to generate egocentric basketball motion sequences in the 12D camera configuration space from a single first-person image. As a result, there are no prior established baselines for this task. To validate our method's effectiveness, we compare it with other popular deep learning techniques such as RNNs~\cite{Jain:1999:RNN:553011}, LSTMs~\cite{Hochreiter:1997:LSM:1246443.1246450}, and GANs~\cite{NIPS2014_5423}. We construct these baselines using the ResNet-101~\cite{DBLP:journals/corr/HeZRS15} architecture. Each baseline takes a single first-person image as input and is then trained to generate $10$ future configurations. The RNN and LSTM baselines are optimized with an L2 loss in the 12D camera configuration space, whereas the GAN baseline has the same architecture as RNN, but is trained to fool the discriminator by producing realistic 12D future camera configurations. To maximize the amount of available training data, the training data are constructed by using every feasible training image as a starting point. During testing, each baseline predicts camera configurations recurrently until we observe no significant change in the prediction or until the predicted sequence length exceeds the length of a real sequence. We also include a nearest neighbor baseline, which operates in the 12D camera configuration space.




The goal of our evaluations is to verify that the generated sequences (1) are realistic and (2) that they capture the goals of real players. To do this, we propose several evaluation schemes, which we describe in the next few subsections.

\subsection{Quantitative Results}

\textbf{Predicting Future Motion.} To verify that our generated sequences are realistic, we examine whether our method can predict a player's future motion. In the context of such evaluation, the sequences that we want to compare typically have different lengths, which renders standard Euclidean and L1 distance metrics unusable. Thus, we use the Hausdorff distance~\cite{Huttenlocher:1993:CIU:628305.628513}, which is commonly used for sequential data comparisons.  To implement this idea, we first use our EgoCam CNN to extract camera configurations from every frame in every real sequence in the testing dataset. Then, as our evaluation metric, we compute a distance $d=D(y,\hat{y})$, where $\hat{y}$ is generated using the first image from a real sequence $y$, and $D$ is a Hausdorff distance operator.  

In Column 2 of Table~\ref{quant_table}, we record the average $d$ over all testing sequences for every method (i.e., the lower the distance the better). We observe that our method outperforms all the other baselines, suggesting that we can use it for future behavior prediction.

\textbf{Capturing the Goals of Real Players.} Earlier, we assumed that the last few frames in real sequences approximate the goals of the players. To examine how well our method captures these goals, we first select the last generated configuration $\hat{y}_M$ in the sequence of length $M$. Then, we use nearest neighbor search to retrieve a real configuration $y_{nn}$ that is most similar to $\hat{y}_M$. Finally, we look up where in the real sequence $y_{nn}$ appears. For instance, if $y_{nn}$ was the $i^{th}$ configuration in a real sequence of length $N$, then we assign our current prediction a value of $v=i/N$. Intuitively, if the last generated configuration is similar to the last real configuration, we conclude that our method was able to capture the goals of real players.

In Column 3 of Table~\ref{quant_table}, we record the average $v$ values over all testing sequences for every method (i.e., the higher the better). These results suggest that, out of all baselines, our method is the most accurate at capturing the goals of real players. Also, note that, if we remove the goal verifier, these results drop significantly, indicating the goal verifier's importance to our system. We conjecture that this happens because the future CNN is trained using a non-discriminative L2 loss, while the goal verifier network is trained to discriminate which configurations represent the final goals of the players. Thus, similar to GANs~\cite{NIPS2014_5423}, due to the discriminative training, the goal verifier may be more useful for generating the configurations that capture the final goals of the players. We also note that the future CNN is still essential, as using a goal verifier network alone produces short and unrealistic sequences. 

\subsection{Qualitative Results}



\textbf{Future CNN Visualization.} To better understand how the future CNN works, we visualize its outputs in Figure~\ref{fut_vis_fig}. To do this, we take the 12D camera configurations generated by the future CNN, transform them into 3D space, and then visualize them in a sparsely reconstructed 3D scene. Note that the future CNN produces a wide array of different configurations, allowing us to generate diverse sequences.

Furthermore, in Columns 2 and 3 of Figure~\ref{cam_seqs_fig}, we visualize the activations of a future CNN from the \textit{res4a} and \textit{res5c} layers, respectively. We observe that, in the \textit{res4a} layer, the CNN focuses on different parts of a defender's body, which makes intuitive sense, as basketball players often use such information to decide their next action. Furthermore, we also observe that, in the \textit{res5c} layer, the future CNN learns to recognize open spaces in the court that could be used by a player to navigate the court and get away from a defender.

\textbf{Visualizing Generated Motion Trajectories.} In the last column of Figure~\ref{cam_seqs_fig}, we also visualize sequences generated by our inverse synthesis procedure (by projecting them onto a 2D court). Even though it is difficult to validate whether our generated motion sequences represent optimal motion pattern, they do seem reasonable. In all visualized instances, our model produces motion sequences that avoid colliding with the defender. Furthermore, most of our generated sequences end around the basket, which reflects what a real player would likely try to do.


\textbf{Retrieved Image Sequences.} To show what our generated sequences look like in the form of first-person images, we use nearest neighbor search in the 12D camera space to retrieve the most similar first-person images from the training data. We then sample $5$ first-person images from the sequence and visualize them in Figure~\ref{img_fig}, along with a real sequence. The similarity between our generated sequences and the real sequences suggests that our model accurately captures how real players move during a basketball game.

\begin{figure}
\begin{center}
    \includegraphics[width=1.02\linewidth]{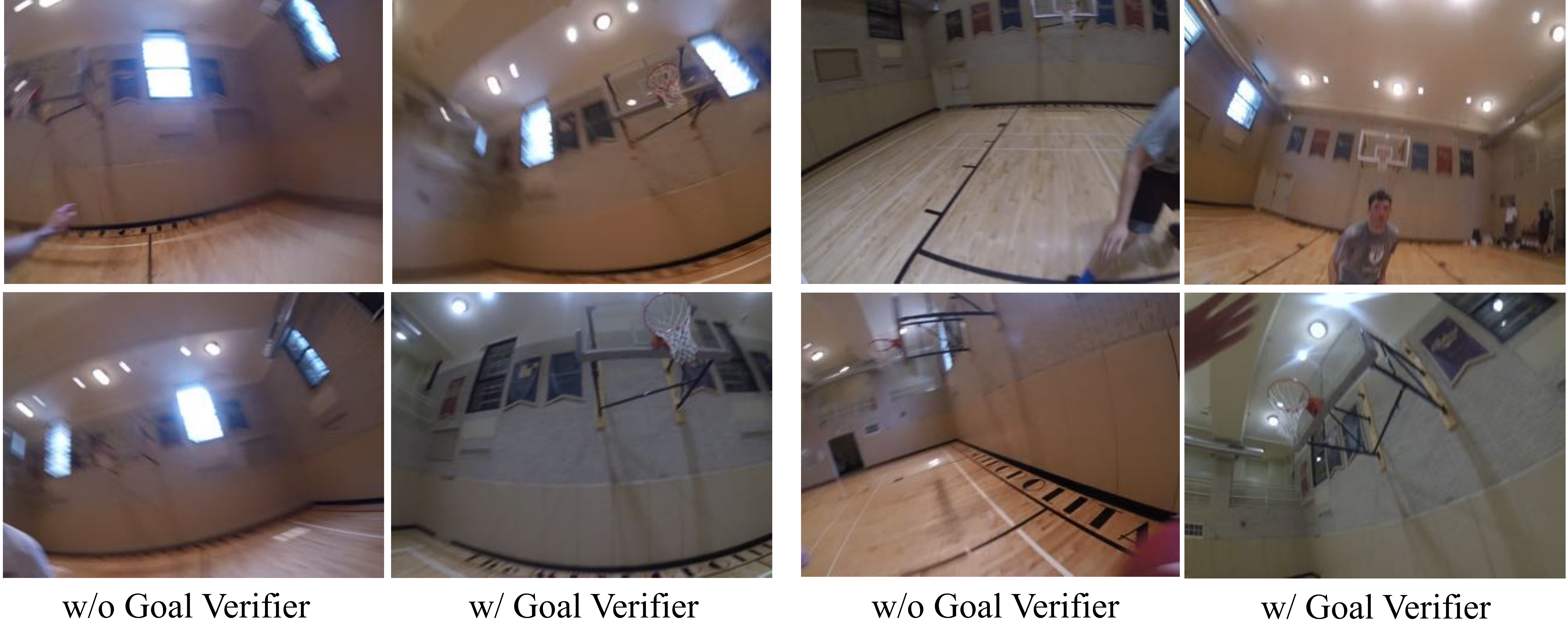}
\end{center}
\vspace{-0.4cm}
   \caption{A figure comparing the final images of our generated sequences in two settings: \textbf{without} and \textbf{with} using the goal verifier network. We retrieve these images via nearest neighbor search. Note that the images produced with a goal verifier ``focus'' on the basket, just as real players would, thus capturing the goals of real players more accurately.\vspace{-0.4cm}}
   
\label{goals_fig}
\end{figure}

Additionally, to qualitatively validate the need for a goal verifier, in Figure~\ref{goals_fig}, we visualize the very last images in our generated sequences, when the goal verifier is not used (w/o Goal Verifier) and when it is used (w/ Goal Verifier). Note that when we remove the goal verifier from our system, the very last images in our generated sequences look less realistic. That is, they focus on the walls or floor of the basketball court, which is probably not something that real players would do. In contrast, adding a goal verifier to our system makes these images focus on the basket, which makes much more sense, since most players finish their sequences when they are looking directly at the basket. Thus, the goal verifier allows us to generate more realistic sequences.



\textbf{Video Results.} Due to space constraints, we cannot include all of our qualitative results in an image format. Furthermore, because images are static, they cannot capture the full content of our generated sequences. Thus, we include more video results in the following link: \url{https://www.youtube.com/watch?v=wRRRl4QsUQg}.

\section{Conclusions}

In this work, we introduced a model that uses a single first-person image to generate an egocentric basketball motion sequence in the form of a 12D camera configuration trajectory. We showed that our model generates realistic sequences that capture the goals of real players. Furthermore, we demonstrated that our model can be learned directly from the first-person video data, which is beneficial as obtaining labeled behavioral data is costly.

Our model could be used for a wide array of behavioral applications, such as future behavior prediction or player development, for which we could build models using the data of expert players and then transfer their behavioral patterns to less-skilled players.  Furthermore, even though we apply our model on a basketball activity, we do not inject any basketball-specific domain knowledge into the model. Thus, we believe that our model is general enough to also be applied to other activities, which we will explore in our future work. In the future, we would also like to experiment with more complex configurations that would allow us to represent even more complicated behavioral patterns. Finally, we believe that the concept behind our model could be used for various robotic applications, where our model could provide behavioral guidelines for robots.

\bibliographystyle{plain}
\footnotesize{
\bibliography{gb_bibliography_v2,bib_hs_v2}}

\end{document}